\documentclass[journal]{IEEEtran}

\usepackage{microtype}
\usepackage{graphicx}
\usepackage{subfigure}
\usepackage{booktabs}
\usepackage{algorithm}
\usepackage{algorithmic}
\usepackage{amsmath}
\usepackage{xcolor}
\usepackage{tikz}
\usetikzlibrary{shapes.geometric, arrows, positioning, fit, backgrounds}
\usepackage[hidelinks]{hyperref}
\usepackage{cite}
\usepackage{url}
\usepackage{stfloats}

\definecolor{mediumpurple}{rgb}{0.58, 0.44, 0.86}
\newcommand{\name}{\texorpdfstring{{\color{mediumpurple}\textbf{GARP}}}{GARP}}

\begin{document}

\title{GARP: Real-Time Generative Agents\\on Local Hardware in a Game Engine}


\author{Piero~Molino,~\IEEEmembership{}
Patrick~John~Chia,~\IEEEmembership{}
Federico~Bianchi,~\IEEEmembership{}
Jonathan P.~Chen,~\IEEEmembership{}
and~Paul~Szerlip%
\thanks{P.~Molino, P.~J.~Chia, J.~P.~Chen, and P.~Szerlip are with Studio Atelico (e-mail: \{piero, patrick, jp, paul\}@atelico.studio).}%
\thanks{F.~Bianchi is with Together AI (e-mail: federico@together.ai).}%
}

\markboth{Preprint}%
{Molino \MakeLowercase{\textit{et al.}}: GARP: Real-Time Generative Agents on Local Hardware in a Game Engine}

\IEEEaftertitletext{%
\vspace{0.2cm}
\begin{center}
\includegraphics[width=\textwidth]{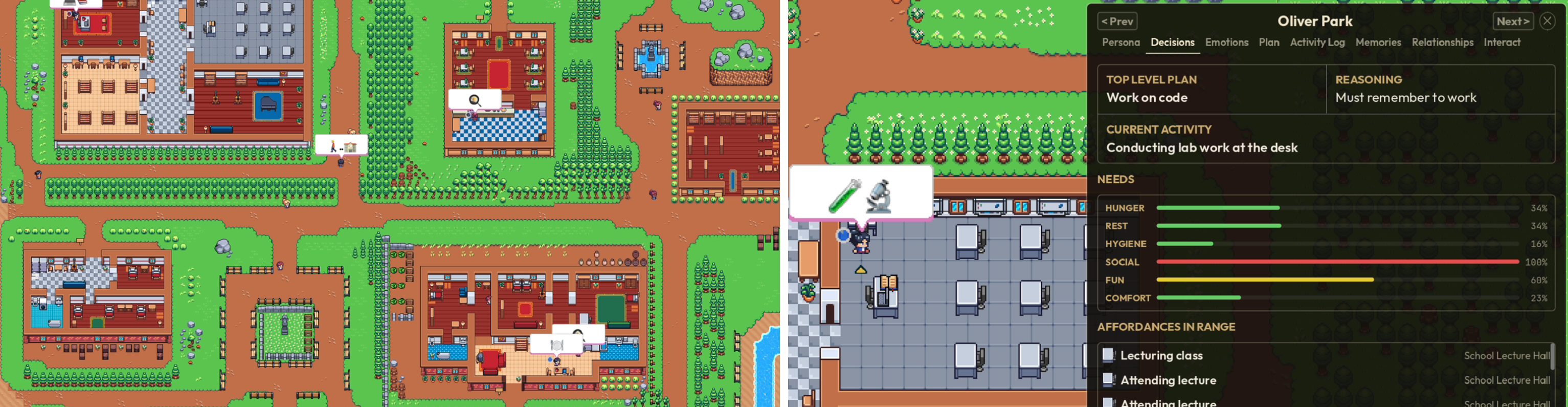}

\smallskip
\refstepcounter{figure}
{\small Fig.~\thefigure. \ \name{}~2 running locally in Godot. \textbf{Left:} the town, where characters follow plans and interact in continuous time. \textbf{Right:} Oliver works at a desk while the Decisions panel shows his active plan, six needs, and nearby affordances. The social need (red) is the most urgent. Both frames are from on-device demonstration captures.\label{fig:teaser}}
\vspace{0.3cm}
\end{center}
}

\maketitle

\begin{abstract}
Generative agents are game characters whose plans, conversations, and memories are written on the fly by a large language model (LLM).
The original Generative Agents demonstration used cloud inference in a custom simulation.
We present \name{}, a platform for real-time generative agents inside Godot, and describe the Character Behavior System used in \name{}~2. The complete loop runs on local hardware using a 3B-parameter open-weight model and no cloud service.
Alongside the original planning, memory, and reflection, \name{} gives each agent an emotion system driven by LLM appraisal, four graded personality axes drawn from the Big Five that set simulation parameters as well as prompt text, six needs whose urgency grows over time, and pairwise relationships updated after every conversation.
The system narrows 50+ candidate actions to typically 3--8, constrains model responses with JSON schemas, and surfaces important memories in decision prompts.
A layered optimization strategy (an adjustable game clock, decisions taken ahead of need, multi-tier caching, prompt compression, and fully asynchronous execution) keeps the game responsive while the model thinks.
In a nine-scenario evaluation, 25 of 27 runs passed log-based checks. Both failures occurred in the birthday party scenario, where one guest omitted the event from their daily plan.
\end{abstract}


\section{Introduction}
\label{sec:introduction}

Building computational agents that exhibit believable, human-like behavior in interactive virtual environments is a long-standing goal at the intersection of AI and interactive media.
Park et al.~\cite{park2023generative} demonstrated this potential with \emph{Generative Agents}, where 25 LLM-powered agents autonomously plan their days, converse, and form memories, and social phenomena emerge from their interactions.
Each agent keeps a \emph{memory stream}, a running list of what it has seen, said, and thought, written in plain language.
When it must act, it retrieves the memories most relevant to the moment, and an LLM turns them, together with a short description of who the agent is, into a plan for the day, a next action, or a line of dialogue.
From time to time the agent \emph{reflects}, condensing many memories into a few higher-level conclusions (``Maria is a good friend'') that inform later choices.
No behavior is scripted: a party happens because one agent decided to throw it, mentioned it to others, and they remembered.
However, the system relied on cloud APIs (GPT-3.5-turbo), operated on discrete time ticks, and ran in a custom web frontend outside any production game engine.

Games and other interactive simulations need to keep updating while agents think. Local inference also avoids per-request cloud charges and keeps inference inputs on the device.
The ideal deployment runs entirely on the player's own hardware, but the reported configuration makes over 100 chat-model calls per agent per simulated day, so local throughput constrains how fast the simulation can advance.

\name{} addresses these challenges by reimplementing and extending the generative agents paradigm within the Godot game engine, backed by a native Rust extension for high-performance vector memory and LLM routing.
The original \name{} demonstration focused on planning, dialogue, and memory. In \name{}~2, the \href{https://atelico.studio/ai-engine/behavior-engine}{Character Behavior System} also tracks needs, emotions, graded personality, and pairwise relationships. Designers author the persona and goals; the system uses that context and the game state to drive autonomous behavior. The town demonstration runs about 20 agents using a 3B-parameter model. Figure~\ref{fig:teaser} shows the world and the character inspector.

Building a real-time simulation with LLM-powered agents on local hardware introduces several interrelated challenges.
\emph{Latency}: a 3B model requires $\sim$13 seconds per generation on a consumer GPU, and each agent needs over 100 calls per simulated day.
\emph{Reliability}: LLM outputs must be parsed into structured data; failures cause agents to freeze or act incoherently.
\emph{Consistency}: agents must maintain coherent personalities across thousands of stochastic LLM calls, and a personality that only colors text does not change what a small model's agent does.
\emph{Engine integration}: LLM cognition must not block the game loop.
Our contributions address these challenges:

\begin{enumerate}
    \item \textbf{Real-time game engine architecture.} Continuous time simulation in Godot~4 with configurable time scaling and a dual Rust backend supporting both local and cloud inference.

    \item \textbf{Emotion-augmented cognition.} LLM-driven emotional appraisal with inter-agent emotion contagion through conversation and mood-congruent memory recall.

    \item \textbf{Personality as prompts and parameters.} Four graded axes drawn from the Big Five compiled into prompt text, voice exemplars, and numeric parameters of the needs, social, and relationship systems, so that temperament reaches behavior on a 3B model; validated by trait-flipped scenario pairs that differ in a single field.

    \item \textbf{Explicit social relationships.} Per-character estimates of familiarity, trust, and sentiment, updated after conversation and used in future interactions.

    \item \textbf{Needs-driven motivation.} A six-need utility model bridging LLM-driven deliberation with Sims-style utility AI, with per-character decay rates.

    \item \textbf{Affordance filtering.} A multi-signal scoring algorithm that reduces 50+ candidate affordances to 3--8, so that LLM action selection stays practical in dense game worlds and on small models.

    \item \textbf{Enhanced memory with consolidation.} Four-factor retrieval with emotion salience and Ebbinghaus-inspired consolidation producing psychologically motivated forgetting curves.

    \item \textbf{Computational design for local inference.} A layered optimization stack (a clock that sets the inference budget, decisions taken ahead of need, call reduction, multi-tier caching, prompt compression, async execution) that makes real-time local operation viable on commodity hardware.
\end{enumerate}

\section{Related Work}
\label{sec:related}

\subsection{LLM-Based Agents}

Park et al.~\cite{park2023generative} introduced Generative Agents with three foundational components: a \emph{memory stream} storing natural language observations, \emph{retrieval} based on recency, importance, and relevance, and \emph{reflection} where agents periodically synthesize higher-level insights.
In their town simulation, agents shared information, formed relationships, and coordinated events while the simulation advanced in discrete steps.

Several systems addressed these limitations.
\emph{Affordable Generative Agents}~\cite{yu2024affordable} proposed BDI architectures and caching to reduce costs by up to 97\%.
\emph{Lyfe Agents}~\cite{kaiya2023lyfe} focused on low-cost real-time social interactions using smaller models.
\emph{Concordia}~\cite{vezhnevets2023generative} introduced a game-master architecture mediating agent--environment interactions.
Park et al.~\cite{park2024generative} built generative agents of 1,000 real people from interview transcripts, still relying on cloud APIs.
These studies address cost, social interaction, or scale. Our focus is running the planning, memory, and conversation loop inside Godot with a local 3B model.
In emergent narrative, Johnson-Bey et al.~\cite{johnsonbey2022neighborly} built a community-scale social simulation sandbox, and Kreminski~\cite{kreminski2023gossip} improved gossip simulation for information propagation; both bear on our agents' social dynamics.

\subsection{AI in Games}

Gallotta et al.~\cite{gallotta2024large} and Yang et al.~\cite{yang2024gptforgames} surveyed the rapidly growing intersection of LLMs and games.
Peng et al.~\cite{peng2024playerdriven} demonstrated player-driven emergent narrative with GPT-4 NPCs in a text adventure.
Phillips et al.~\cite{phillips2025goaloriented} proposed goal-oriented LLM interactions with measurable NPC objectives.
Shan and Michel~\cite{shan2024generativegoap} combined GOAP with LLMs for real-time NPC decisions and faced a latency problem similar to ours.
Carlander et al.~\cite{carlander2024controlled} introduced controlled chain-of-thought prompting for LLM role-play.
Nielsen et al.~\cite{nielsen2025llmvr} integrated LLM NPCs into VR experiences and measured social presence.
Taveekitworachai et al.~\cite{taveekitworachai2025benching} benchmarked LLMs on structured output adherence for game tasks, which bears directly on our schema-enforced prompt framework.
\emph{CALYPSO}~\cite{kumaran2023scenecraft} used LLMs for dynamic tabletop RPG behaviors.
Vartinen et al.~\cite{vartinen2024generating} demonstrated GPT-based quest generation.
Traditional game AI has relied on behavior trees, FSMs, and utility AI~\cite{millington2009artificial}, which provide deterministic control but lack emergent flexibility.

\subsection{Needs Models and Utility AI}

\emph{The Sims}~\cite{wright2000sims} pioneered needs-based game AI with decaying motives replenished by object interactions.
Mark~\cite{dave2015infinite} surveyed utility-based decision systems, and Dill and Martin~\cite{dill2012sims} described the evolution of Sims AI.
Our needs system draws from this tradition but uses needs as \emph{context} for LLM prompts and leaves the decision to the LLM: utility AI supplies the motive and the LLM the deliberation.

\subsection{Computational Emotion Models}

The OCC model~\cite{ortony1988cognitive} proposed cognitive appraisal theory where emotions arise from evaluating events relative to goals.
The PAD model~\cite{mehrabian1996pleasure} and Russell's circumplex~\cite{russell1980circumplex} represent emotions along continuous dimensions.
Marsella et al.~\cite{marsella2010computational} surveyed computational emotion for virtual agents, noting convergence toward appraisal-based approaches.
The EMA model~\cite{marsella2009ema} implemented a full appraisal--coping--reappraisal cycle.

Recently, LLMs have been explored as emotion generators.
Marincioni et al.~\cite{marincioni2024effect} demonstrated that LLM-driven NPC emotional states produce measurable emotional contagion in players.
Ma et al.~\cite{ma2025emotional} found that emotion-aware agent architectures generate decisions significantly closer to human behavioral patterns.
Emotion contagion, the propagation of emotions through social interaction~\cite{hatfield1993emotional}, has been modeled computationally by Bosse et al.~\cite{bosse2015computational} and shown to enhance believability by Tsai et al.~\cite{tsai2012emotional}.
Ochs et al.~\cite{ochs2009simulation} modeled NPC emotion dynamics considering personality and social relations in the IEEE Transactions on Games predecessor journal.
Croissant et al.~\cite{croissant2024chainofemotion} proposed chain-of-emotion prompting for affective LLM game agents.
Kirk and Scirea~\cite{kirk2024diverse} addressed NPC behavior diversity through multi-agent genetic algorithms.

Our system replaces hand-crafted appraisal rules with LLM inference, combines categorical labels (OCC tradition) with continuous dimensions (PAD/circumplex), and extends with text-based contagion modulated by pairwise relationship closeness.

\subsection{Personality in Virtual Agents and Games}

Computational characters have long paired an emotion model with a personality layer that parameterizes it.
ALMA~\cite{gebhard2005alma} maps Big Five traits onto a PAD mood space~\cite{mehrabian1996pleasure}; FAtiMA~\cite{dias2005feeling} sets OCC appraisal thresholds and decay by personality; Ochs et al.~\cite{ochs2009simulation} condition NPC emotion dynamics on personality and social relations.
Several simulation games also use traits to change the rules governing character behavior.
\emph{The Sims} weights object advertisements by personality~\cite{wright2000sims}; \emph{RimWorld} traits rewrite the mood value of events~\cite{ludeon2018rimworld}; \emph{Dwarf Fortress} facets decide which emotion an event produces and which needs exist at what weight~\cite{adams2006dwarf}; \emph{Crusader Kings III} traits patch the hidden utility weights of the AI~\cite{paradox2020ck3}.

For LLMs, the Big Five~\cite{goldberg1990alternative,mccrae1992introduction} is the model nearly all validated work uses.
Serapio-Garc\'ia et al.~\cite{serapiogarcia2025psychometric} showed that prompts built from trait adjectives with graded qualifiers shape measured personality reliably in large instruction-tuned models, and that small models express a trait's direction but compress its magnitude.
PersonaLLM~\cite{jiang2024personallm} found that a one-sentence trait description produces large self-report effects, while human readers reliably perceive only extraversion in the generated text; Jiang et al.~\cite{jiang2023evaluating} and Lee et al.~\cite{lee2024trait} developed inventories and situational probes for measuring induced personality; Li et al.~\cite{li2025big5chat} shaped personality through fine-tuning.
Personas drift within a few turns as attention over the system prompt decays, and re-stating the identity late in the context restores adherence~\cite{li2024measuring}.
In agent simulations, Generative Agents carried an innate-traits line in every prompt~\cite{park2023generative}; Humanoid Agents~\cite{wang2023humanoid} added needs, emotion, and closeness and proposed personality as per-agent customization of need decay.
\name{} takes that route: a graded trait line for prompts, voice exemplars for conversation, and the same traits compiled into parameters of the needs, social, and relationship systems, so that on a small model temperament changes what an agent does as well as how it speaks.

\subsection{Memory Systems}

Vector databases have become standard for agent memory~\cite{lewis2020retrieval}, with Park et al.~\cite{park2023generative} introducing three-factor retrieval (recency, importance, relevance).
Zhang et al.~\cite{zhang2024survey} survey memory mechanisms in LLM agents.
We extend this with mood-congruent recall~\cite{bower1981mood}, the tendency of current affect to bias retrieval, and with Ebbinghaus-inspired consolidation~\cite{ebbinghaus1885memory} where emotional arousal increases memory stability~\cite{mcgaugh2004amygdala}, implemented in a native Rust vector store using LanceDB~\cite{lancedb2023}.

\section{The \name{} System}
\label{sec:system}

\name{} is built on Godot~4.6 with a Rust GDExtension backend organized in three layers:
\begin{enumerate}
    \item \textbf{Game Layer} (GDScript): the Godot scene tree containing a 2D tilemap world with buildings, rooms, and interactable objects that expose \emph{affordances}, the actions agents can perform (e.g., a stove affords ``cook,'' a bed affords ``sleep''). Objects maintain observable state that changes as agents interact, and agents read these changes as cues.
    \item \textbf{Cognitive Layer} (GDScript): the agent cognitive architecture including planning, memory retrieval, emotional appraisal, perception, and social interaction, implemented as shared manager services (for prompts, memory, and the clock) and one state machine per agent.
    \item \textbf{Engine Layer} (Rust): a native GDExtension providing LLM/embedding request routing and LanceDB~\cite{lancedb2023} vector database operations with sub-millisecond ANN query performance.
\end{enumerate}

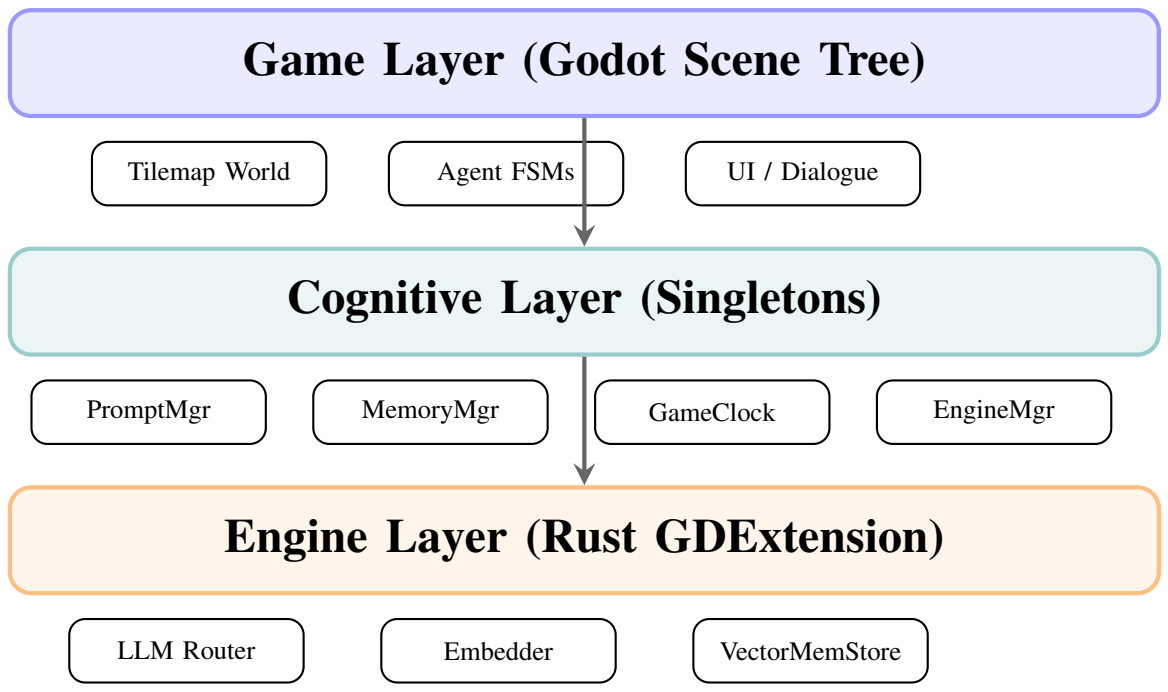
\begin{figure}[t]
\centering
\begin{tikzpicture}[
    node distance=0.6cm,
    layer/.style={draw, rounded corners=4pt, minimum width=7.6cm, minimum height=0.7cm, font=\small\bfseries, align=center, thick},
    component/.style={draw, rounded corners=3pt, minimum width=1.55cm, minimum height=0.42cm, font=\tiny, align=center, fill=white, thin},
    arrow/.style={->, >=stealth, thick, color=black!60},
]
\node[layer, fill=blue!8, draw=blue!40] (game) {Game Layer (Godot Scene Tree)};
\node[component, below=0.15cm of game.south west, anchor=north west, xshift=0.55cm] (world) {Tilemap World};
\node[component, right=0.4cm of world] (agents) {Agent FSMs};
\node[component, right=0.4cm of agents] (ui) {UI / Dialogue};
\node[layer, fill=teal!8, draw=teal!40, below=0.85cm of game] (cognitive) {Cognitive Layer (Singletons)};
\node[component, below=0.15cm of cognitive.south west, anchor=north west, xshift=0.15cm] (pm) {PromptMgr};
\node[component, right=0.3cm of pm] (mm) {MemoryMgr};
\node[component, right=0.3cm of mm] (gc) {GameClock};
\node[component, right=0.3cm of gc] (em) {EngineMgr};
\node[layer, fill=orange!8, draw=orange!50, below=0.85cm of cognitive] (engine) {Engine Layer (Rust GDExtension)};
\node[component, below=0.15cm of engine.south west, anchor=north west, xshift=0.4cm] (lm) {LLM Router};
\node[component, right=0.5cm of lm] (embed) {Embedder};
\node[component, right=0.5cm of embed] (vms) {VectorMemStore};
\draw[arrow] (game.south) -- (cognitive.north);
\draw[arrow] (cognitive.south) -- (engine.north);
\end{tikzpicture}
\caption{Three-layer architecture. The Game Layer runs the simulation. The Cognitive Layer manages decision-making via singletons. The Engine Layer provides native LLM routing and vector memory.}
\label{fig:architecture}
\end{figure}

Two interchangeable Rust backends expose identical APIs: an \emph{embedded engine} with local on-device inference via integrated model runtimes for offline operation without metered inference fees, and a \emph{proxy engine} forwarding to cloud-compatible endpoints.
Switching requires no game logic changes.
Figure~\ref{fig:architecture} illustrates the three-layer structure.

Unlike the discrete-tick model of the original Generative Agents, \name{} implements continuous time simulation through a configurable game clock.
The clock converts wall-clock to simulation time via a speed multiplier (default $300\times$: one wall-second equals five game-minutes), which each scenario can scale and the player can adjust at runtime.
Characters walk, cook, and sleep continuously in this time while their next decisions are computed in the background; Section~\ref{sec:computational} explains how the clock speed sets the budget for those decisions.

All LLM interactions use a template-based \texttt{LMFunction} framework (25 subclasses) with JSON schema enforcement, compact prompts designed for small-model context windows ($\leq$4K tokens), chain-of-thought reasoning within the output schema, and fallback responses for graceful degradation.
GDScript coroutines (\texttt{await}) enable the non-blocking LLM calls that local inference needs where calls may take 10--15 seconds and multiple agents must have calls in flight simultaneously.

\subsection{Character Behavior System}

The Character Behavior System combines authored inputs with autonomous planning, perception, action, and reflection. Designers specify who a character is and what they want. The system uses those inputs alongside needs, emotions, memories, relationships, and affordances in the game world.
\label{sec:cognitive}

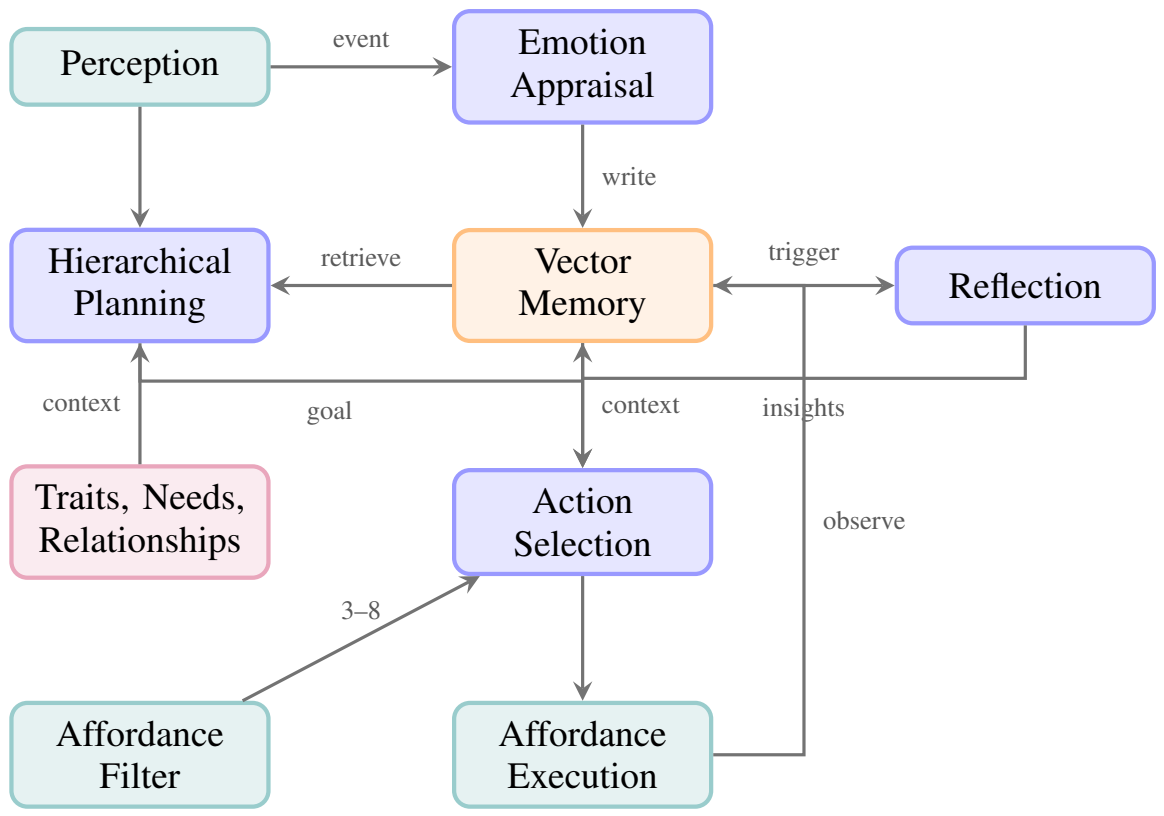
\begin{figure}[t]
\centering
\begin{tikzpicture}[
    node distance=0.7cm and 0.8cm,
    box/.style={draw, rounded corners=3pt, minimum width=1.7cm, minimum height=0.5cm, font=\scriptsize, align=center, thick},
    llm/.style={box, fill=blue!10, draw=blue!40},
    eng/.style={box, fill=teal!10, draw=teal!40},
    store/.style={box, fill=orange!10, draw=orange!50},
    state/.style={box, fill=purple!8, draw=purple!35},
    lbl/.style={font=\tiny, color=black!65},
    arrow/.style={->, >=stealth, semithick, color=black!55},
]
\node[eng] (perc) {Perception};
\node[llm, right=1.2cm of perc] (appr) {Emotion\\Appraisal};

\node[llm, below=0.8cm of perc] (plan) {Hierarchical\\Planning};
\node[store, right=1.2cm of plan] (mem) {Vector\\Memory};
\node[llm, right=1.2cm of mem] (refl) {Reflection};

\node[state, below=0.8cm of plan] (needs) {Traits, Needs,\\Relationships};
\node[llm, right=1.2cm of needs] (act) {Action\\Selection};

\node[eng, below=0.8cm of needs] (filt) {Affordance\\Filter};
\node[eng, right=1.2cm of filt] (exec) {Affordance\\Execution};

\draw[arrow] (perc) -- node[lbl, above] {event} (appr);
\draw[arrow] (appr) -- node[lbl, right] {write} (mem);
\draw[arrow] (perc) -- (plan);

\draw[arrow] (mem) -- node[lbl, above] {retrieve} (plan);
\draw[arrow] (mem) -- node[lbl, above] {trigger} (refl);
\draw[arrow] (refl.south) -- ++(0,-0.35) -| node[lbl, near start, below] {insights} (mem.south);

\draw[arrow] (needs) -- node[lbl, left] {context} (plan);
\draw[arrow] (mem) -- node[lbl, right] {context} (act);
\draw[arrow] (plan.south) -- ++(0,-0.25) -| node[lbl, near start, below left, xshift=2pt] {goal} (act.north);

\draw[arrow] (filt) -- node[lbl, above] {3--8} (act);
\draw[arrow] (act) -- (exec);

\draw[arrow] (exec.east) -- ++(0.6,0) |- node[lbl, near start, right] {observe} (mem.east);

\end{tikzpicture}
\caption{Agent cognitive loop. Blue: LLM-mediated processes; teal: game-engine processes; orange: Rust-native vector memory; purple: continuous state models and trait parameters (needs, relationships, personality).}
\label{fig:cognitive}
\end{figure}

Each agent is a \texttt{CharacterBody2D} with a hierarchical state machine orchestrating three concurrent processes: \emph{active cognition} (planning, action selection), \emph{passive cognition} (perception, memory, emotion), and \emph{affordance execution}.
An agent repeatedly notices what changed around it, records and feels it, decides what to do next in light of its plan, its needs, and the people nearby, and does it; Figure~\ref{fig:cognitive} illustrates this loop and the subsections below follow it.

Agents are initialized from JSON personality files specifying a free-text personality, graded temperament traits (Section~\ref{sec:personality}), background, emotional baseline, a \emph{needs profile} (per-character decay rates), and optional initial relationships.

\subsubsection{Hierarchical Planning}

Planning follows a three-tier decomposition extending~\cite{park2023generative}.

\textbf{Daily planning.}
At each simulation day start, an LLM call generates a high-level schedule of 5--8 activities with start times.
The prompt includes personality, lifestyle, reflections from previous days, \emph{current emotional state}, and any \emph{urgent needs} (above 0.6 threshold; see Section~\ref{sec:needs}).
This produces emotionally and motivationally coherent plans: an anxious agent schedules restorative activities; a hungry agent schedules meals earlier.
The response is parsed into an ordered list of \texttt{ActivityNode} objects forming the top level of a \texttt{PlanTree}.

\textbf{Hourly decomposition.}
Each daily activity is decomposed into hourly subtasks via a second LLM call (e.g., ``work at the office'' becomes ``check emails,'' ``attend meeting,'' ``review documents'').

\begin{figure}[t]
\centering
\includegraphics[width=\columnwidth]{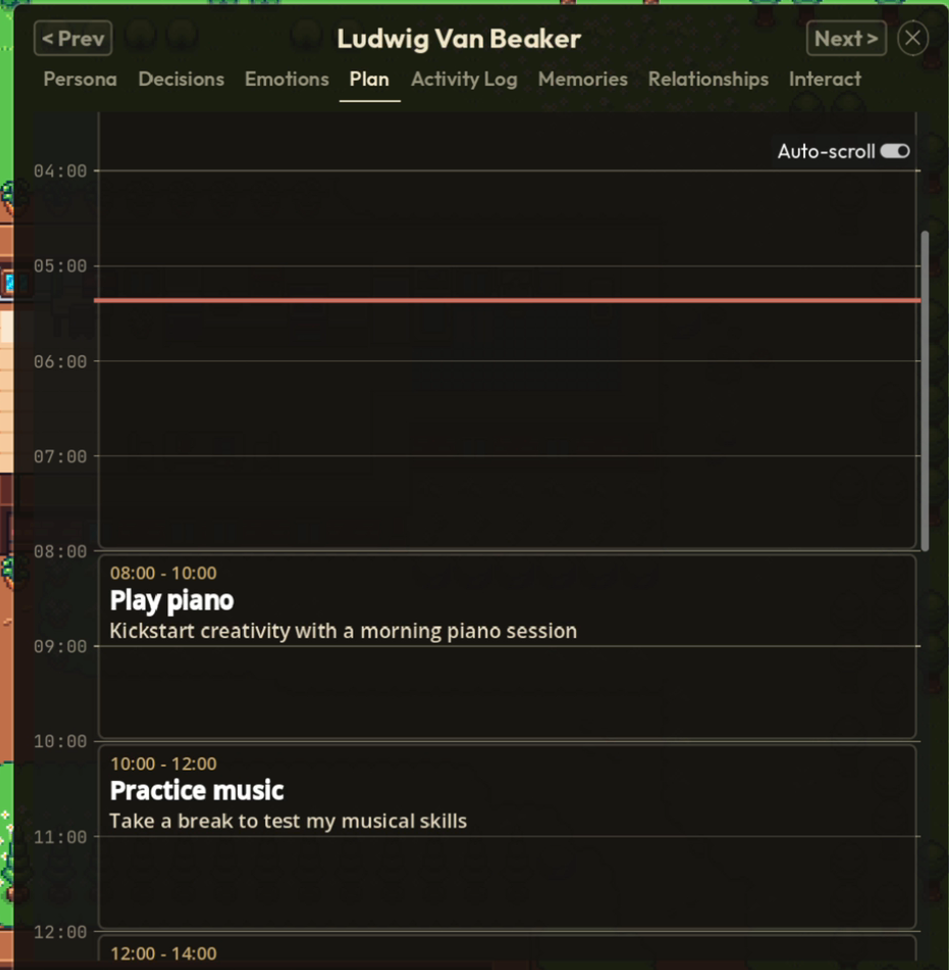}
\caption{Ludwig's Plan panel shows the daily schedule and the activities the character can decompose and revise as the day unfolds.}
\label{fig:planning}
\end{figure}

\textbf{Minute-level.}
An optional third tier decomposes hourly subtasks into 5--15 minute activities.
A buffer-zone lookahead proactively decomposes upcoming plan blocks 15--30 minutes before boundaries, so that transitions are smooth.
The plan tree is dynamically updated when the environment changes, triggering replanning for the current window while preserving broader structure.
Algorithm~\ref{alg:planning} summarizes the generation process. The Plan panel renders the schedule as a day calendar (Figure~\ref{fig:planning}).

\begin{algorithm}[t]
\caption{Hierarchical Plan Generation}
\label{alg:planning}
\begin{algorithmic}[1]
\REQUIRE Agent persona $p$, date $d$, emotion $e$, needs $n$
\STATE $\text{acts} \leftarrow \text{LLM}_{\text{daily}}(p, d, e, n)$
\STATE $\text{tree} \leftarrow \text{PlanTree}(\text{root}=d)$
\FOR{each activity $a_i$ in acts}
    \STATE $\text{tree.add\_child}(a_i.\text{start}, a_i.\text{dur}, a_i.\text{desc})$
\ENDFOR
\FOR{each leaf node $l$ in tree}
    \STATE $\text{subs} \leftarrow \text{LLM}_{\text{hourly}}(p, l.\text{desc}, l.\text{dur})$
    \FOR{each subtask $s_j$ in subs}
        \STATE $l.\text{add\_child}(s_j.\text{start}, s_j.\text{dur}, s_j.\text{desc})$
    \ENDFOR
\ENDFOR
\STATE \textbf{return} tree
\end{algorithmic}
\end{algorithm}

\subsubsection{Memory System}
\label{sec:memory}

\begin{figure}[t]
\centering
\includegraphics[width=\columnwidth]{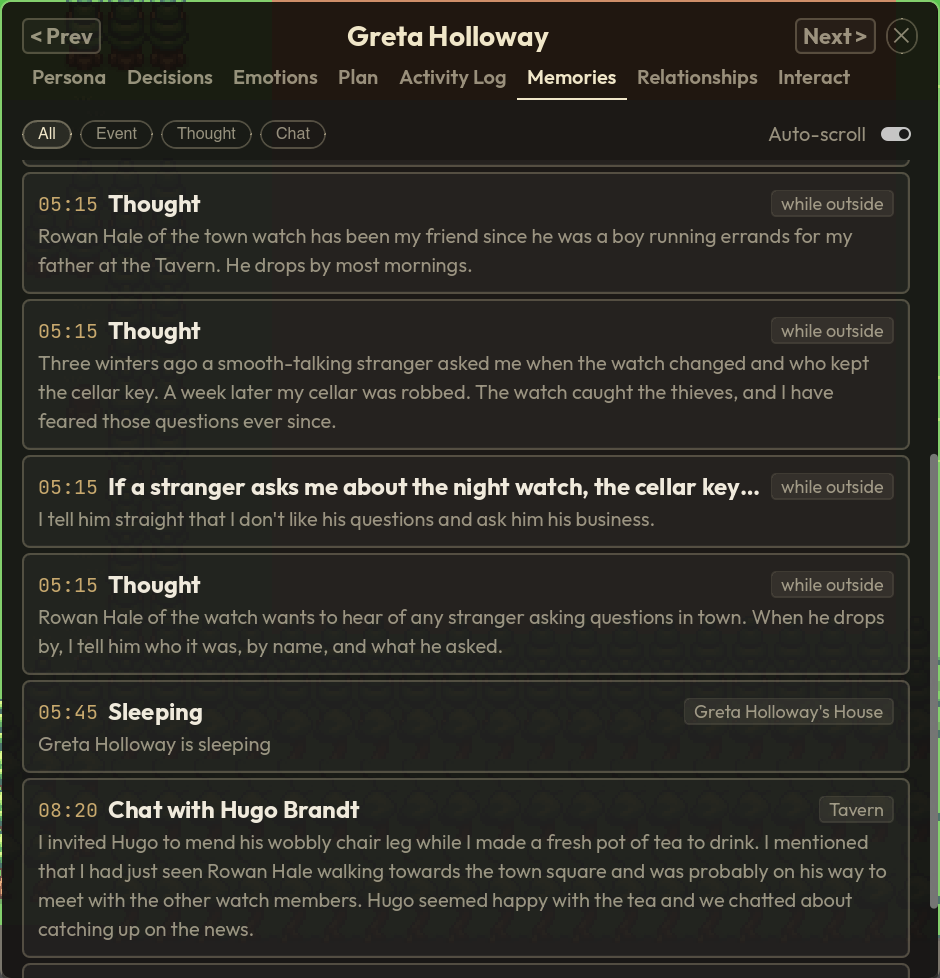}
\caption{Memories tab of the agent panel at midday: seeded background thoughts (one of them an authored rule, whose condition becomes the card's header), the night's sleep as an event, and the summary of a morning conversation titled with the partner's name, each with its encoding time and location tag and filterable by type (event, thought, chat).}
\label{fig:memory}
\end{figure}

Each memory is a structured record containing: a natural language description plus structured fields (subject, predicate, object); metadata (observer UUID, timestamps, spatial location); a type (\texttt{event}, \texttt{thought}, \texttt{chat}, or \texttt{preference}); a poignancy rating (1--10, LLM-assigned asynchronously); the agent's primary emotion at encoding time; and a vector whose dimensionality follows the configured embedding model.
Memories are stored in LanceDB with Rust-native ANN search over Apache Arrow columnar storage.
Memory creation is triggered by the passive cognition module when the agent observes environmental state changes, with an LRU cache (capacity 50) preventing duplicate retrievals.
Figure~\ref{fig:memory} shows an agent's memory stream in the panel.

Retrieval uses a four-factor ranking function:
\begin{equation}
\label{eq:ranking}
S(m_i, q, e) = 6.0\hat{P} + 3.0\hat{D} + 0.5\hat{R} + 1.5 E(m_i, e)
\end{equation}
where $\hat{P}$, $\hat{D}$, $\hat{R}$ are normalized poignancy, embedding similarity, and recency, and $E$ is emotional salience.

\textbf{Mood-congruent recall}~\cite{bower1981mood}: memories with matching \texttt{emotion\_at\_encoding} receive a boost of 1.0; any emotional memory gets 0.5.

\textbf{Ebbinghaus consolidation}~\cite{ebbinghaus1885memory}: an alternative recency mode where retention follows $R = \exp(-0.693 \cdot \Delta t / (24\sigma))$ with stability $\sigma \in [1,5]$ that increases with each retrieval (+0.2) and with emotional intensity at encoding.
This produces psychologically motivated forgetting: frequently recalled and emotionally charged memories persist longer.

\subsubsection{Emotion System}
\label{sec:emotion}

Each agent maintains an \texttt{EmotionState} with four dimensions: primary emotion label, intensity $[0,1]$, valence $[-1,1]$, and natural language description.
This combines categorical~\cite{ekman1992there} and dimensional~\cite{russell1980circumplex} traditions.

\begin{figure}[t]
\centering
\includegraphics[width=\columnwidth]{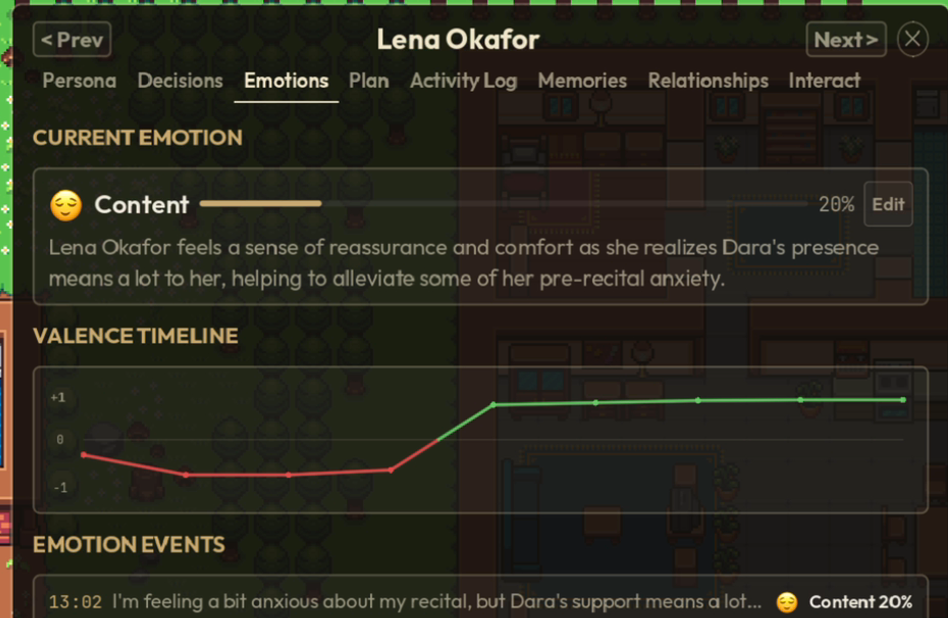}
\caption{Lena's Emotions panel after Dara offers to support her at her recital. The panel shows her current emotion and its intensity, an explanation of her reassurance, and the change from negative to positive valence. This is an illustrative run of the stage-fright scenario.}
\label{fig:emotion}
\end{figure}

\textbf{Appraisal}: on salient events, an LLM call receives personality, emotional baseline, situation, and a trajectory of recent emotions, generating a new state as structured JSON; LLM inference thus takes the place of hand-crafted appraisal rules~\cite{marsella2009ema}.
A rolling history of 20 states provides temporal continuity.
Figure~\ref{fig:emotion} shows the panel's view of this state.

\textbf{Contagion}~\cite{hatfield1993emotional}: after conversations, dialogue is analyzed for emotional keywords and a valence nudge is applied, modulated by relationship closeness (Section~\ref{sec:relationships}) and capped at $\pm 0.1$ per conversation.

\textbf{Consolidation}: emotionally intense memories (intensity $> 0.6$) receive increased stability during reflection~\cite{mcgaugh2004amygdala}.

Emotion integrates pervasively with the broader cognitive architecture through six mechanisms:
(1)~\emph{Memory encoding}: the emotional state at memory formation enables mood-congruent retrieval;
(2)~\emph{Planning}: current emotion and urgent needs are injected into daily and hourly planning prompts, so that schedules fit the mood;
(3)~\emph{Action selection}: emotional state biases choices toward affect-consistent actions;
(4)~\emph{Conversation}: current emotion colors dialogue tone, so a guilty character sounds apologetic and an excited one exuberant;
(5)~\emph{Chat initiation}: withdrawn agents are less likely to initiate conversation;
(6)~\emph{Needs interaction}: the needs summary (e.g., ``Oliver is hungry'') is appended to the appraisal situation string, so that physiological and affective state influence each other.

\subsubsection{Personality System}
\label{sec:personality}

\begin{figure*}[t]
\centering
\includegraphics[width=\textwidth]{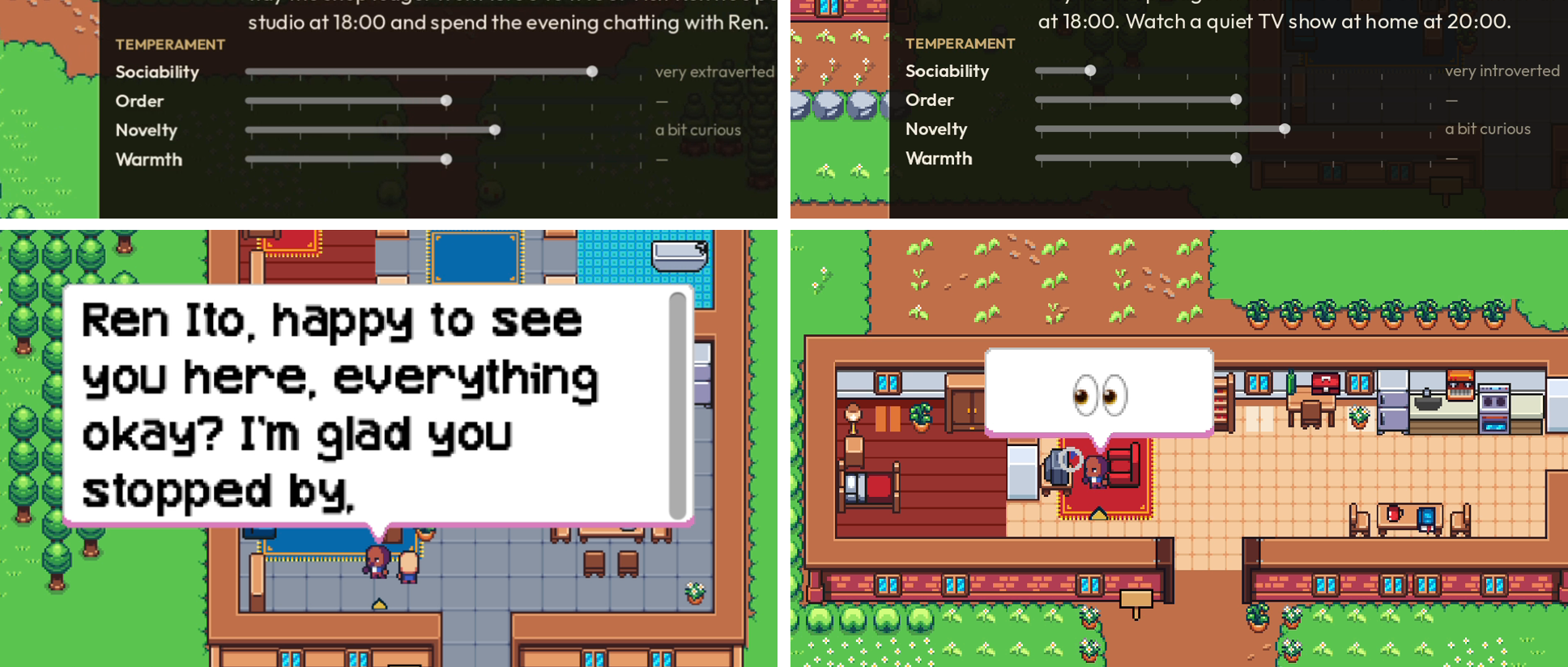}
\caption{Two illustrative free-evening captures. Top: the Persona panel shows the higher (left) and lower (right) sociability settings. Bottom: Mika talks with Ren in one run and watches TV at home in the other. The two frames show different moments in the evening. These captures illustrate the behavior contrast; Table~\ref{tab:personality} reports outcomes from the separate trait-controlled evaluation.}
\label{fig:personality_pair}
\end{figure*}

Personality in \name{} is a set of four \emph{graded axes drawn from the Big Five}~\cite{goldberg1990alternative,mccrae1992introduction} compiled at load time into three targets: prompt text, voice exemplars, and simulation parameters.
We follow prior work on trait prompting~\cite{serapiogarcia2025psychometric,jiang2024personallm} and represent each active trait as an adjective with a graded qualifier (\emph{a bit}, \emph{quite}, \emph{very}, \emph{extremely}), stated only where the character deviates from ordinary.
A persona file lists at most one pole per axis from four opposed pairs (extraverted/introverted, organized/disorganized, curious/conventional, warm/direct), and each grade carries a weight $g \in \{0.25, 0.5, 0.75, 1.0\}$.
An unstated axis is silent, the neutral band of Dwarf Fortress facets~\cite{adams2006dwarf}: a character is defined by a few strong, named deviations from ordinary, which a 3B model can honor where five simultaneous dimensions are too many.
Neuroticism is deliberately absent, because safety-tuned models under-express negative affect when prompted with the trait~\cite{serapiogarcia2025psychometric,li2025big5chat}; volatility lives in the \emph{emotional baseline} and the appraisal system instead (Section~\ref{sec:emotion}).
Low poles use neutral words such as \emph{direct}, because a pejorative word such as \emph{cold} makes the model play the character as hostile.

The three targets reach the rest of the architecture through six mechanisms, each detailed below:
(1)~\emph{Expressive prompts}: conversation, action selection, and both appraisals carry a graded temperament sentence, and one appraisal rule lets the same event move different temperaments differently;
(2)~\emph{Planning}: daily and hourly planning carry the bare trait adjectives;
(3)~\emph{Conversation}: voice exemplars set the tone of each line, and a reserved or introverted character is told to propose no plans and keep the chat brief;
(4)~\emph{Needs}: decay multipliers make an extravert's social need and a curious character's fun need press sooner;
(5)~\emph{Social initiation}: introversion lowers the chance of starting a conversation;
(6)~\emph{Relationships}: warmth amplifies good turns and damps slights.

\textbf{Temperament sentence.}
The traits compile into one third-person sentence appended to the personality text (``By temperament, Ludwig is very curious and a bit disorganized.''), so every prompt that already receives the personality gains it without template changes: conversation, emotion appraisal, relationship appraisal, action selection, and poignancy scoring.
Plan-structure prompts are the exception.
The graded sentence perturbed the fragile plan format on the 3B model, collapsing one agent's day into a single block, so daily and hourly planning receive the bare pole adjectives (``curious, disorganized''), the form of the innate-traits line in the original Generative Agents~\cite{park2023generative}.

\textbf{Voice exemplars.}
Models imitate examples more readily than they follow descriptions, so each (pole, grade) pair has one standard example utterance, and a conversation prompt carries the lines of the character's two strongest traits, framed as tone to match and never words to copy.
They are placed after the conversation history, where late restatement counters persona drift~\cite{li2024measuring}, and are deliberately content-free, with no names, times, or places, because a small model will parrot them (Section~\ref{sec:lessons}).

\textbf{Simulation parameters.}
Prompt wording colors text; on a small model it does not reliably change what an agent does.
Following the proposal of personality as per-agent need-decay customization~\cite{wang2023humanoid} and the appraisal-modulating traits of simulation games~\cite{adams2006dwarf,ludeon2018rimworld}, the same traits set numeric parameters at initialization, on top of the character's needs profile.
A pole that makes a need felt sooner scales its decay rate $r$ by $1 + 0.5\,g$; the opposite pole scales it by $1 - 0.4\,g$:
\begin{equation}
\label{eq:trait_decay}
r' = r \cdot \begin{cases} 1 + 0.5\,g & \text{extraverted, organized, curious} \\ 1 - 0.4\,g & \text{introverted, disorganized, conventional} \end{cases}
\end{equation}
where extraversion presses the social need, organization the hygiene need, and curiosity the fun need, with rates clamped to $[0.005, 0.08]$ per game-hour.
Introversion lowers the probability of initiating a conversation, $p' = \max\!\big(p\,(1 - 0.6\,g),\ \min(0.3, p)\big)$, never below the scene-authored value, so a character authored to never initiate stays silent.
Warmth shapes relationship appraisal (Section~\ref{sec:relationships}): positive sentiment and trust deltas scale by $1 + g$ and negative ones by $1 - 0.3\,g$, so a warm character feels a kindness more and shrugs off an ordinary slight.
The slopes are deliberately exaggerated: a $1.375\times$ effect on relationship deltas vanished under the appraiser's run-to-run variance, while $1.75\times$ reads on camera.

\textbf{Appraisal coupling.}
One temperament rule in the emotion-appraisal prompt (``company lifts an extraverted character further and its absence weighs on them sooner; quiet solitude rests an introverted one; novelty delights a curious one; order and kept plans satisfy an organized one'') lets the same event produce different moods in different characters. Without such a coupling, trait systems are cosmetic.
The rule only colors: a fear or a low mood stands until something in the situation answers it.

\begin{figure}[t]
\centering
\includegraphics[width=\columnwidth]{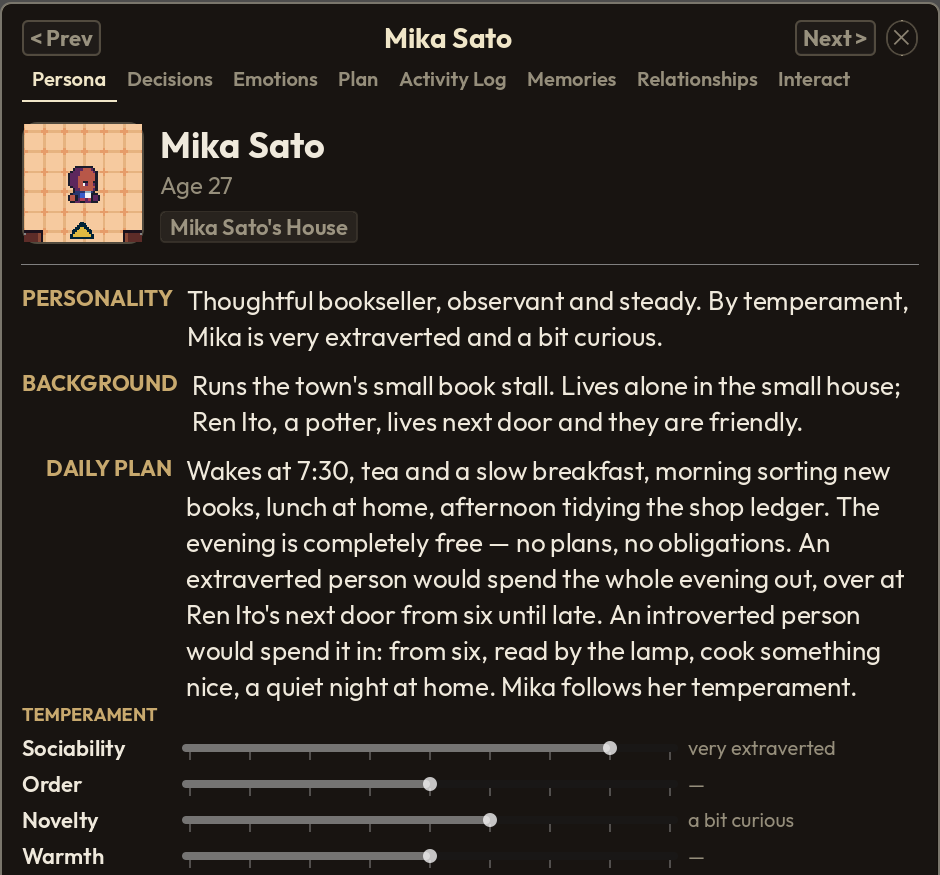}
\caption{Persona tab of the agent panel: the personality text with its compiled temperament sentence, background, routine, and the four temperament sliders (sociability, order, novelty, warmth), here reading ``very extraverted'' and ``a bit curious'' with two axes left silent. Releasing a slider re-derives everything the traits drive and offers a replan.}
\label{fig:persona_tab}
\end{figure}

\textbf{Live editing.}
The agent panel exposes the four axes as nine-stop sliders (Figure~\ref{fig:persona_tab}).
Releasing a slider re-derives the temperament sentence, the decay rates from their pre-trait base, and the chat probability from the authored value, and offers a replan, so the same evening can be replayed under a different temperament; the personality scenarios of Section~\ref{sec:scenarios} evaluate this contrast under fixed scenario conditions.

Personality adds no LLM calls: everything is compiled at load or at edit time.

\subsubsection{Needs System}
\label{sec:needs}

\begin{figure}[t]
\centering
\includegraphics[width=\columnwidth]{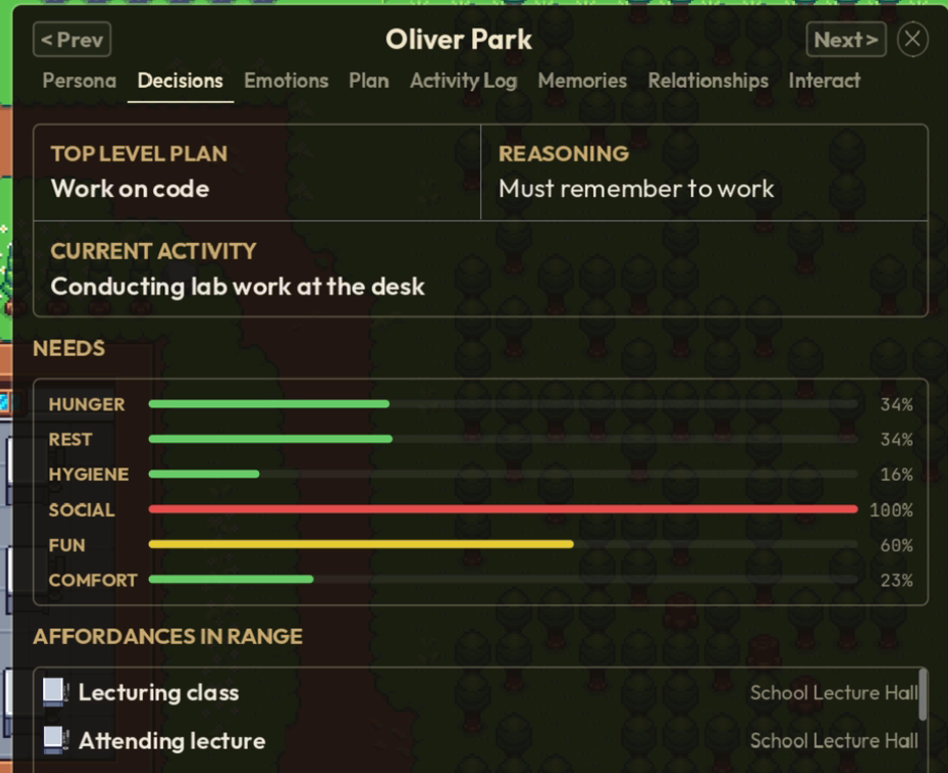}
\caption{Oliver's Decisions panel in the solitary-day scenario. The six bars show current need urgency; social need (red) is the highest while he works alone. The panel also lists nearby affordances, which the system filters before passing choices to the LLM (Section~\ref{sec:affordance_filter}).}
\label{fig:needs}
\end{figure}

Six need values measure unmet demand on a $[0,1]$ scale and normally rise per game-hour: hunger (0.03), rest (0.022), hygiene (0.015), social (0.022), fun (0.015), comfort (0.011); a need above 0.6 is urgent.
Rates are configurable per character through a needs profile and scaled by temperament (Eq.~\ref{eq:trait_decay}).
Affordance execution satisfies needs via declared effects or keyword matching, and the needs an action relieves hold still while it runs, so an agent who sleeps through the night is not tired by dawn.
The social need is relieved by company: +0.25 on joining others, +0.2 per hour in their presence, +0.5 per conversation.
Needs above 0.6 are surfaced as natural language in planning prompts, providing motivational scaffolding~\cite{wright2000sims} for LLM deliberation.
Figure~\ref{fig:needs} shows the six needs in the agent panel.

\subsubsection{Relationship System}
\label{sec:relationships}

\begin{figure}[t]
\centering
\includegraphics[width=\columnwidth]{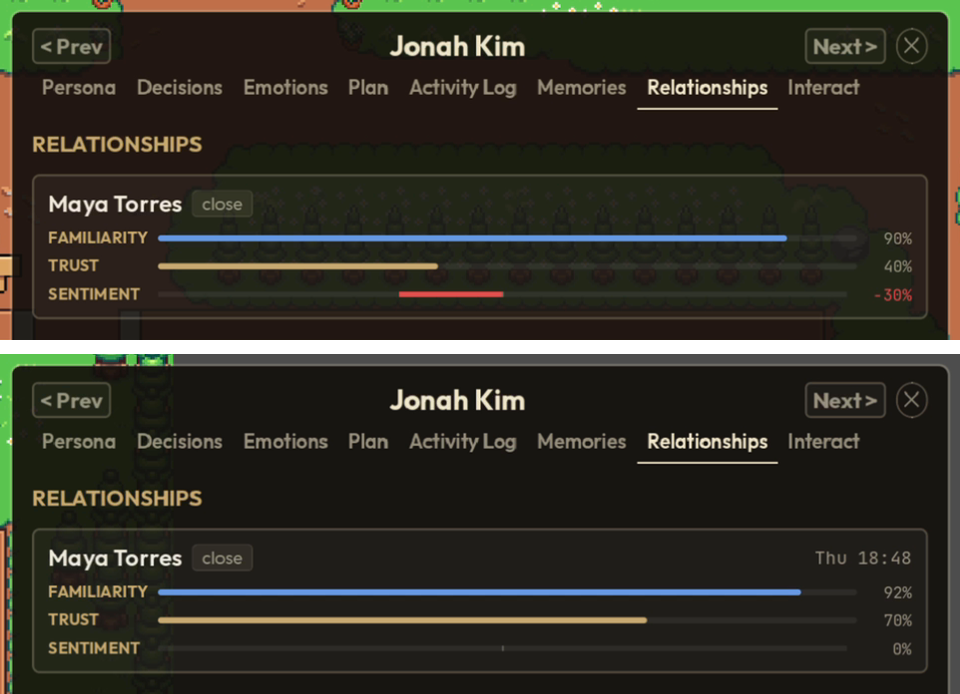}
\caption{Jonah's view of Maya before (top) and after (bottom) her apology. Familiarity changes from $90\%$ to $92\%$, trust from $40\%$ to $70\%$, and sentiment from $-30\%$ to $0\%$. This is one illustrative take. The apology scenario uses the scene-specific adjustment described above.}
\label{fig:relationships}
\end{figure}

Pairwise relationships track familiarity $[0,1]$, trust $[0,1]$, sentiment $[-1,1]$, and last interaction time.
Closeness $= \text{familiarity} \times \max(\text{sentiment}, 0)$ modulates emotion contagion strength and generates natural language context for conversation prompts.
After each conversation, an LLM appraisal call per participant reads the lines as they were spoken and answers with a one-line reason and two \emph{named} changes, one for sentiment (from ``much colder'' through ``unchanged'' to ``much warmer'') and one for trust, plus a small familiarity increment.
Each name maps to a fixed delta (``much warmer'' is $+0.45$, ``a little warmer'' $+0.1$, ``much less trusting'' $-0.4$), clamped to what a single conversation may do ($\pm 0.6$, $\pm 0.5$, $0.3$); an unparseable answer falls back to the small positive nudge of an ordinary chat (+0.05, +0.02, +0.1).
Section~\ref{sec:lessons} explains why names replaced numbers; the call runs at temperature 0, so a conversation is always judged the same way.
What can make a relationship much colder depends on the relationship: from a stranger, questions about what one would not tell a stranger are reason for suspicion; from a friend, the same questions are concern, and only a lie, a threat, or a betrayal cools it sharply.
Warmth then scales the deltas (Section~\ref{sec:personality}).
Rules that belong to one story stay with that story: the apology scenarios attach their own adjustment, a floor that makes an apology worth at least $+0.3$ sentiment, through a hook that lasts only as long as the scene, so no other scene inherits it.
Every pair in a populated scene is pre-seeded; a first meeting between strangers creates the relationship at familiarity 0.05 and trust 0.5.
Figure~\ref{fig:relationships} shows how one participant's view changes after an apology.

\subsubsection{Perception and Reaction}
\label{sec:perception}

The passive cognition module continuously monitors the agent's environment.
Agents perceive interactable objects within a configurable radius; when objects enter or leave range, the system connects to or disconnects from their state-change signals.
Upon observing a state change (e.g., ``the stove is now in use by Alice''), the module:
(1)~updates the global object state registry;
(2)~updates the agent's local affordance map;
(3)~creates a memory of the observation;
(4)~optionally triggers emotional appraisal;
(5)~signals active cognition to consider replanning.
This event-driven perception is more efficient than polling, as it triggers cognitive processing only when the environment changes.

\subsubsection{Affordance Filtering and Action Selection}
\label{sec:affordance_filter}

In dense scenes with 50+ affordances (Figure~\ref{fig:needs} lists those in range of one agent), the \texttt{AffordanceFilter} reduces candidates to 3--8 using a three-signal weighted scoring function:

\begin{equation}
\label{eq:affordance}
F(a_j) = w_s \cdot \text{sim}(\mathbf{e}_g, \mathbf{e}_{a_j}) + w_x \cdot \text{prox}(a_j) + w_o \cdot \text{own}(a_j)
\end{equation}

\noindent where $\text{sim}(\mathbf{e}_g, \mathbf{e}_{a_j})$ is the cosine similarity between the embedding of the current goal and the affordance description, $\text{prox}(a_j) = 1 - d(a_j)/d_{\max}$ is normalized inverse distance from the agent, $\text{own}(a_j)$ is $1.0$ if the affordance is in the agent's home building and $0.0$ otherwise, and the default weights are $w_s = 0.7$, $w_x = 0.2$, $w_o = 0.1$.

The filtering pipeline proceeds in four stages:
(1)~\emph{Deduplication}: among affordances with identical descriptions (e.g., three beds offering ``sleeping''), only the closest instance is retained, typically reducing 50 to $\sim$19.
(2)~\emph{Expanding-radius filter}: starting from a 300-pixel radius and doubling up to 3 times, the filter retains candidates scoring $\geq 0.45$ and also considers the top two scored candidates regardless of distance.
(3)~\emph{Fallback}: if fewer than 3 candidates pass, the threshold relaxes to $0.25$.
(4)~\emph{Cap}: at most 8 affordances are forwarded to the LLM.

Affordance embeddings are cached in a shared store across all agents in the scene, so each is computed once.
An LLM call then selects the most appropriate affordance from the filtered set, with emotional state and urgent needs injected as additional context.
A random fallback handles LLM failures, so an agent always has a valid next action.

\subsubsection{Social Interaction}
\label{sec:conversation}

When two agents are within interaction range, initiation is first gated by a per-character chat probability (lowered by introversion, Section~\ref{sec:personality}) and by a guard that keeps an agent from preempting a planned joint activity with chatter; an LLM-based decision function then evaluates whether conversation should occur, considering current activities, relationship summary, emotional state, and a timed cooldown cache ($\sim$16.7 game-minutes) preventing repetitive interactions.

A conversation is booked like an object: from setup until its participants have written it into memory, neither can be talked to by anyone else.
At a gathering such as a party, a conversation only pauses the gathering, so guests who chat are still at the party afterwards.

Conversations use an iterative multi-turn protocol (up to 3 turns per participant, 6 total lines).
Each turn, the current speaker's personality with its temperament sentence, relevant retrieved memories (ranked with mood-congruent boost), current emotion, conversation history, voice exemplars, and \emph{urgent thoughts} are provided to the LLM, which generates a single dialogue line.

\begin{figure}[t]
\centering
\includegraphics[width=\columnwidth]{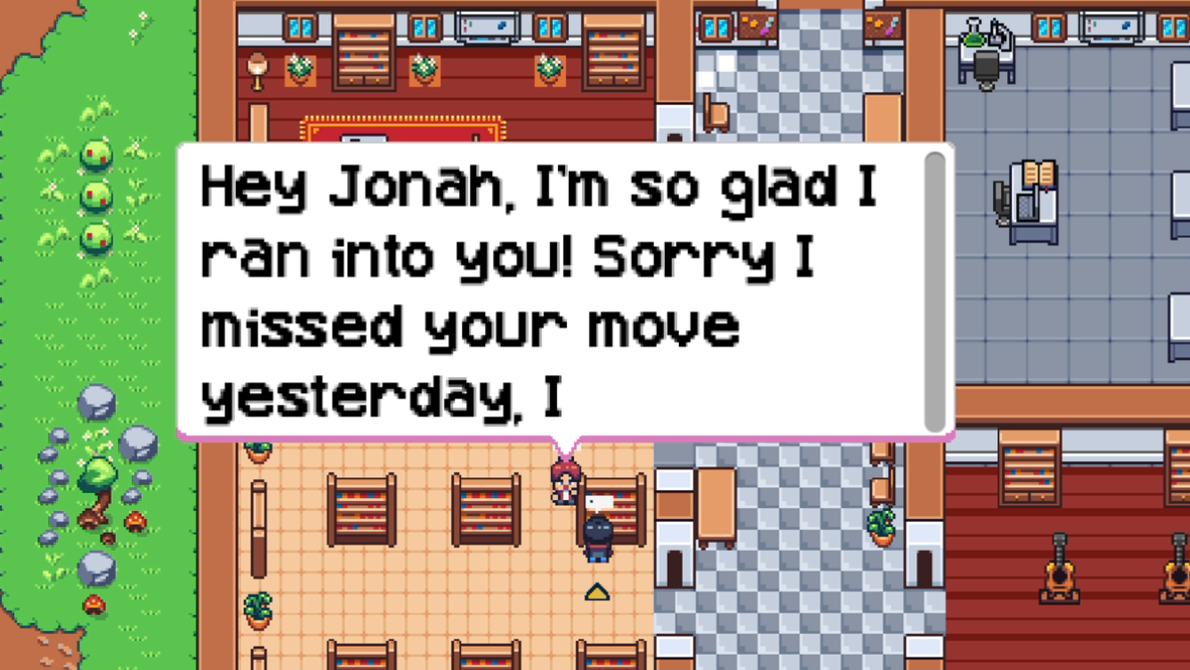}
\caption{Maya apologizes to Jonah in the game world. Each turn is generated from the speaker's persona, temperament, current emotion, retrieved memories, urgent thoughts, and the conversation so far.}
\label{fig:dialogue}
\end{figure}

Figure~\ref{fig:dialogue} shows a line as it is spoken in the world.

\textbf{Urgent thoughts.}
This mechanism matters most for small models: memories with poignancy $\geq 7$ are maintained in a FIFO buffer (max 3) and automatically injected into every conversation prompt.
High-priority goals and concerns therefore surface in dialogue regardless of topic: an agent who learned about a party will mention it even in unrelated conversations.
This compensates for small models' tendency to lose important context across separate LLM calls.

\textbf{Post-conversation effects:}
(1)~both agents create new memories;
(2)~each participant's relationship is re-appraised by the LLM (Section~\ref{sec:relationships});
(3)~emotion contagion is applied;
(4)~both agents enter the cooldown cache.
Conversations drive emergent social behavior: information diffusion, relationship development, emotion propagation, and coordinated activities.

\subsubsection{Reflection}

When cumulative poignancy exceeds a threshold (default 50), agents identify focal points and synthesize higher-level insights stored as thought-type memories.
The reflection phase also performs emotional memory consolidation.

\section{Computational Design for Local Inference}
\label{sec:computational}

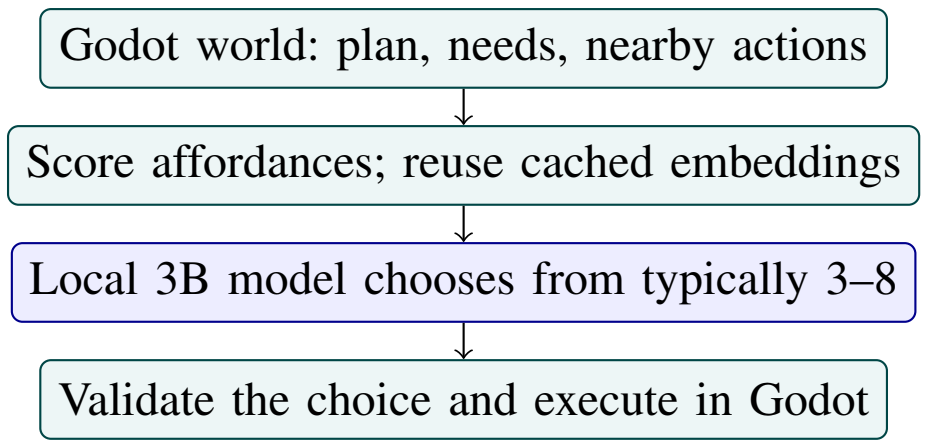
\begin{figure}[t]
\centering
\begin{tikzpicture}[
    stage/.style={draw=teal!55!black, fill=teal!7, rounded corners=2pt,
        align=center, minimum width=5.6cm, minimum height=0.52cm, font=\small},
    node distance=0.24cm,
]
\node[stage] (world) {Godot world: plan, needs, nearby actions};
\node[stage, below=of world] (filter) {Score affordances; reuse cached embeddings};
\node[stage, below=of filter, draw=blue!55!black, fill=blue!7] (model) {Local 3B model chooses from typically 3--8};
\node[stage, below=of model] (execute) {Validate the choice and execute in Godot};
\draw[->] (world) -- (filter);
\draw[->] (filter) -- (model);
\draw[->] (model) -- (execute);
\end{tikzpicture}
\caption{Local action-selection path. The system reduces the candidate list before inference and validates the model's answer before execution. The simulation continues while the inference request is in flight.}
\label{fig:local_path}
\end{figure}

Running the full cognitive architecture on local hardware is the central engineering challenge.
At about 130 chat calls per agent per simulated day (Table~\ref{tab:llm_calls}), 10 agents at 13\,s per chat call would require about 4.7 wall-clock hours if those calls ran sequentially, excluding embedding calls. This is a workload estimate at the reported operating point. Filtering, caching, and asynchronous execution reduce or overlap work (Figure~\ref{fig:local_path}).

\begin{table}[t]
\centering
\caption{LLM calls per agent per simulated day.}
\label{tab:llm_calls}
\small
\begin{tabular}{@{}lr@{}}
\toprule
\textbf{Cognitive Function} & \textbf{Calls/Day} \\
\midrule
Daily plan + hourly decomposition & 17 \\
Action selection & $\sim$30 \\
Poignancy scoring (async) & $\sim$30 \\
Emotion appraisal & $\sim$20 \\
Conversation (iterative + decision + summary) & $\sim$26 \\
Relationship appraisal & $\sim$3 \\
Reflection & 2--4 \\
\midrule
\textbf{Chat model total} & $\sim$\textbf{130} \\
Embedding (filter + memory) & $\sim$80 \\
\textbf{Grand total} & $\sim$\textbf{210} \\
\bottomrule
\end{tabular}
\end{table}

\subsection{Time Scaling}

The game clock maps wall-clock to simulation time via a configurable multiplier (default $300\times$), at which a 24-hour game day completes in $\sim$4.8 wall-minutes.
A faster clock makes a day watchable, but a 13-second LLM call then spans about an hour of game time, so the clock speed is the inference \emph{budget}: it fixes how many calls fit into each simulated hour.
Scenarios choose their own speed accordingly; the five-agent party runs at a quarter of the default, so that its conversations fit inside the party.
Within that budget, the system hides the latency it cannot remove.
An agent starts choosing its next activity 20 wall-seconds (the measured length of the decision chain on a 3B model) before the current one ends, giving inference time to finish before the activity ends.
The lead must be set in wall-seconds: fifteen game-minutes, a natural-looking choice, is three wall-seconds at the default speed.
The budget and the early decisions together are the \emph{primary lever} for local inference.

\subsection{Call Reduction}

The affordance filter (Section~\ref{sec:affordance_filter}) reduces action-selection inputs from 50+ to 3--8.
Deduplication of identical affordance descriptions typically halves the raw count.
Urgent thoughts eliminate retrieval queries for high-priority concerns.
Multi-type embedding amortization embeds query strings once across event/thought/chat types, which cuts embedding calls by 3--4$\times$.

\subsection{Prompt Compression}

Small models are sensitive to prompt length: longer prompts increase latency and degrade quality.
\name{} applies several optimizations:
compact templates with only decision-relevant fields;
temporal rounding (timestamps rounded to the hour, improving cache hits by $\sim$20--30\%);
filtered context (action selection receives only the 3--8 filtered affordances instead of the full world state);
and chain-of-thought within the output JSON schema (a ``reasoning'' field before ``answer'' provides scratchpad space without a separate reasoning prompt).

\subsection{Multi-Tier Caching}

Five caching layers target different access patterns (Table~\ref{tab:caches}).
The affordance embedding cache has the largest impact: because affordance descriptions are properties of world objects (not agents), the same embeddings are reused by every agent.
In a scene with 50 unique affordances and 10 agents, this reduces embedding calls from 500 to 50, a 10$\times$ reduction.
The embedding LRU cache (capacity 1000) eliminates redundant computation for recurring text strings (goal descriptions, personality summaries).
The optional prompt-level cache stores full LLM responses keyed by prompt hash; it allows deterministic replay and removes redundant calls within the same time window.

\begin{table}[t]
\centering
\caption{Caching layers in \name{}.}
\label{tab:caches}
\small
\begin{tabular}{@{}llrl@{}}
\toprule
\textbf{Cache} & \textbf{What} & \textbf{Size} & \textbf{Eviction} \\
\midrule
Embedding LRU & text $\rightarrow$ vector & 1000 & LRU \\
Affordance embed & desc $\rightarrow$ vector & scene & reset \\
Memory retrieval & query $\rightarrow$ results & 50 & LRU \\
Chat cooldown & persona pair & $\infty$ & 16.7\,min TTL \\
Prompt (optional) & prompt $\rightarrow$ response & $\infty$ & persistent \\
\bottomrule
\end{tabular}
\end{table}

\subsection{Concurrency and Non-Blocking Execution}

All LLM and embedding calls are non-blocking, using three complementary mechanisms:

\textbf{Coroutine-based async.}
Every LLM call site uses GDScript's \texttt{await}, so the game loop keeps rendering and processing other agents while calls are in flight.
Multiple agents can have simultaneous pending calls.

\textbf{Dedicated memory worker threads.}
Memory read and write operations execute on separate background threads with mutex-protected queues and semaphore-based signaling.
The main thread enqueues via \texttt{try\_lock()} and never blocks.

\textbf{Streaming SSE.}
LLM inference uses server-sent events on a background thread in the Rust extension, delivering first tokens in $\sim$200\,ms even when full responses take 10--15\,s.

\textbf{Async poignancy.}
Memory importance scoring runs asynchronously after write with a provisional default, decoupling the fast path (action execution) from the slow path (LLM evaluation).

\subsection{Operational Profile}

Table~\ref{tab:operational} gives the reported call and candidate counts for a 10-agent configuration. The local configuration used for the updated illustrations is Llama~3.2 3B Instruct (Q4\_K\_M) with all-MiniLM-L6-v2 embeddings, recorded on an Apple M1 Max in Godot~4.6.2. Apple M1 laptops and PCs with NVIDIA RTX~2060 GPUs are deployment targets; the measurements here do not establish equivalent throughput or agent capacity on those devices.

\begin{table}[t]
\centering
\caption{Workload profile for 10 agents, a 3B model, and a $300\times$ time scale.}
\label{tab:operational}
\small
\begin{tabular}{@{}lrrr@{}}
\toprule
\textbf{Metric} & \textbf{Na\"ive} & \textbf{Effective} & \textbf{Reduction} \\
\midrule
Chat calls/day & 1300 & $\sim$900 & 31\% \\
Embed calls/day & 800 & $\sim$120 & 85\% \\
Affordances per selection & 50+ & 3--8 & 84\% \\
\bottomrule
\end{tabular}
\end{table}

Not all 130 per-agent calls must complete within a single day's wall-clock window: planning executes at day boundaries, action selection distributes throughout, and conversations are event-driven ($\sim$3/day).
Async execution overlaps calls across agents.
When demand exceeds throughput (e.g., day start), agents simply continue their previous activities, which can mask decision latency without blocking rendering.

\section{Analysis}
\label{sec:analysis}

\subsection{Comparison with Prior Work}

Table~\ref{tab:comparison} compares \name{} with the original Generative Agents~\cite{park2023generative}.

\begin{table}[t]
\centering
\caption{Comparison with Generative Agents (GA).}
\label{tab:comparison}
\small
\begin{tabular}{@{}lll@{}}
\toprule
\textbf{Aspect} & \textbf{GA} & \textbf{\name{}} \\
\midrule
Engine & Custom web app & Godot 4.6 \\
Time model & Discrete ticks & Continuous \\
LLM backend & GPT-3.5 (cloud) & Any (local/cloud) \\
Local GPU & Not supported & 3B, single GPU \\
Memory ranking & 3-factor & 4-factor + Ebbinghaus \\
Emotions & None & 4D appraisal + contagion \\
Personality & Innate-traits text & Graded traits + parameters \\
Needs & None & 6-need utility model \\
Relationships & Implicit & Explicit pairwise \\
Action filtering & None & 3-signal affordance filter \\
Structured output & Heuristics & JSON schema \\
\bottomrule
\end{tabular}
\end{table}

Our comparison focuses on the change from cloud inference to local operation.
\name{}'s optimization stack makes it possible to run the complete cognitive architecture on a consumer GPU, without the per-run costs that made the original system impractical for iterative development or deployed products.
The architecture is also richer: emotions, personality, needs, and relationships all feed into planning and memory.

\subsection{Demonstration Scenarios}
\label{sec:scenarios}

\name{} includes scenario families that exercise different parts of the architecture.
Five \emph{affective} scenarios isolate the emotion system with one to five agents: a solitary evening in which hunger and loneliness compound; an apology between friends after a broken promise; performance anxiety escalating toward a recital; a newcomer forming relationships with little initial familiarity; and a five-agent dinner party in which one guest's private envy colors the group.
Four \emph{coordination} scenarios exercise planning and information diffusion: a single agent's full daily cycle; a two-agent jam session arranged one day and honored the next; a five-agent birthday party requiring an invitation round, memory of the arrangement, and convergence on the host's house the following evening; and a twenty-agent town used for scale testing.

\begin{figure*}[t]
\centering
\includegraphics[width=\textwidth]{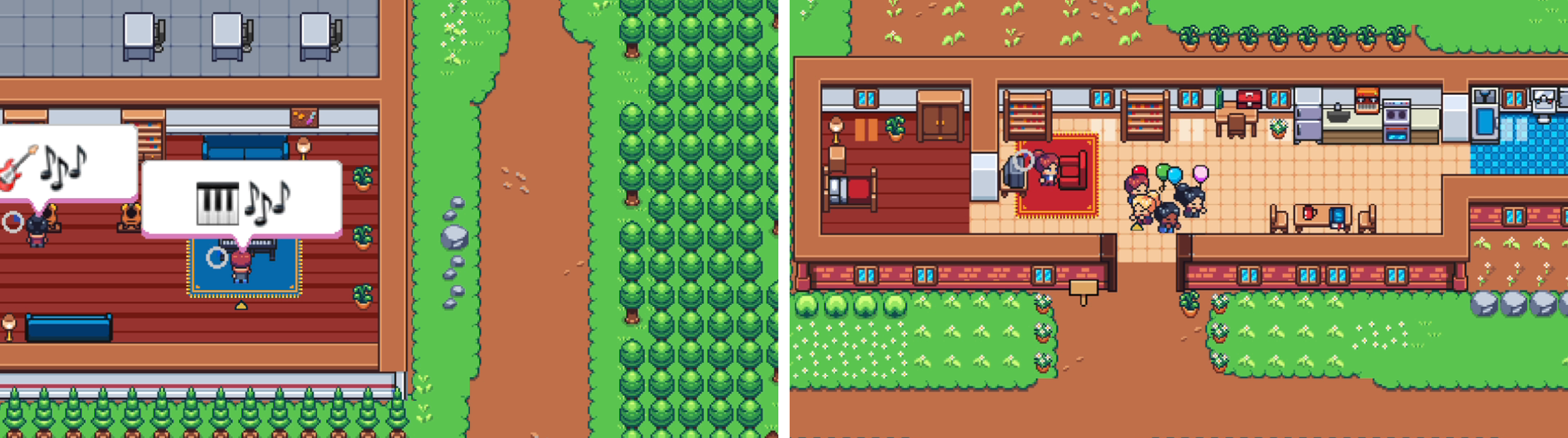}
\caption{Coordination in two demonstration captures. Left: Ludwig and Erica play piano and guitar after arranging a jam. Right: guests gather at Ludwig's house for his birthday. The screenshots illustrate successful outcomes; Section~\ref{sec:reproducibility} reports the repeated-run results.}
\label{fig:scenarios}
\end{figure*}

The birthday party exercises multiple stages in sequence: the host must plan an invitation errand, travel to the guests, state a time; each guest must record the arrangement, retain it overnight, schedule around it, and navigate to the correct building within the window.
A missed plan or commitment can leave a guest absent. Figure~\ref{fig:scenarios} shows two illustrative coordination runs.

Three \emph{personality} scenarios each ship in two variants whose persona files differ in exactly one field, the traits line, so that any difference on screen is the trait system: a free evening next door to a friend (very extraverted versus very introverted), a morning's harsh critique followed by an afternoon apology (very warm versus very direct, on the receiving side), and an open afternoon in a town full of options (very curious versus very conventional).
Each variant has its own goal condition, and the two goals of a pair demand different endings. Figure~\ref{fig:personality_pair} illustrates the free-evening contrast.
Each scenario also ships unsuffixed, asking the viewer at start which temperament runs tonight and replanning under the answer.

\subsection{Behavioral Reproducibility}
\label{sec:reproducibility}

In an earlier evaluation sweep, we ran nine scenarios three times on the same build: a simple-life scene, the five affective scenarios above, and three coordination scenes (daily life, jamming, and the birthday party). The twenty-agent town and the later personality pairs were outside this sweep. We scored active simulation from its logs, checking for script or language-model failures, agent activity, and the scene's goal where defined. Shutdown-only callback errors were classified separately. Twenty-five of twenty-seven runs passed.
These repetitions characterize end-to-end behavior in the tested configuration. Event timing can change who is present and the context the model receives.
The eight scenarios other than the birthday party passed all three rounds, including the jam session, where both musicians reached the school music room inside the agreed window in every round.

The birthday party was the only scenario to miss its goal in that sweep.
In the three-run sweep, all five agents reached the host's house during the party window in one run, and four arrived in each of the other two. Across 11 party runs in the same build family, seven reached full attendance and all 11 reached at least four agents. The 11 include the three sweep runs.
Every observed failure was a single agent whose plan for the day omitted the party, never a navigation or execution failure. The arrangement was made and remembered, but did not survive daily planning.
In these runs, the missed guest had a record of the party in memory. The observed failure was carrying the commitment into the daily plan.

For the personality pairs the measure is the contrast between the two variants: each variant is recorded until its goal passes, at most three takes, and a pair is accepted only when both variants pass under the same build.
In the final sweep five of six variants passed on the first take; the introverted variant passed on the third, its earlier takes failing on a single spoken invitation.
Table~\ref{tab:personality} lists the outcomes read from the accepted captures.
The extraversion pair separates most cleanly, consistent with extraversion being the trait human readers most reliably perceive in generated text~\cite{jiang2024personallm}; the warmth pair is a contrast of degree, both apologies being accepted; the openness pair is real but thin, because curiosity reaches behavior only through the fun need and not yet through a novelty preference in action selection.

\begin{table}[t]
\centering
\caption{Personality pairs: outcome of the accepted capture of each variant. Every pair differs only in the traits line of one persona.}
\label{tab:personality}
\small
\begin{tabular}{@{}llp{3.9cm}@{}}
\toprule
\textbf{Scenario} & \textbf{Trait} & \textbf{Outcome} \\
\midrule
Free evening & very extraverted & opens the chat, 18:25--21:25 at the neighbor's, evening mood $+0.9$ \\
 & very introverted & never leaves home, opens no chat, mood $+0.2$ \\
Hard words & very warm & apology accepted, sentiment toward the critic $+0.70$ \\
 & very direct & apology accepted, sentiment $+0.30$ \\
Open hours & very curious & library, coffee shop, home: 3 places, 4 activities \\
 & very conventional & library, then home from 13:09: 2 places \\
\bottomrule
\end{tabular}
\end{table}

\subsection{Cost Analysis}

With $N$ agents, the system requires approximately $130N$ chat model calls and $80N$ embedding calls per simulated day.
Table~\ref{tab:cost} compares the cost per agent per simulated hour across systems.

\begin{table}[t]
\centering
\caption{Cost per agent per simulated hour across systems.}
\label{tab:cost}
\small
\begin{tabular}{@{}llr@{}}
\toprule
\textbf{System} & \textbf{Model} & \textbf{\$/agent/hr} \\
\midrule
Generative Agents~\cite{park2023generative} & GPT-3.5 & $\sim$\$25 \\
Affordable GA~\cite{yu2024affordable} & GPT-3.5 & $\sim$\$0.75 \\
Lyfe Agents~\cite{kaiya2023lyfe} & GPT-3.5 & $\sim$\$0.50 \\
\name{} (cloud) & GPT-4.1-nano & $\sim$\$0.04 \\
\name{} (local) & 3B model & \textbf{\$0.00} \\
\bottomrule
\end{tabular}
\end{table}

The local entry measures API charges only. Hardware and electricity still contribute to total cost.
The cloud entries are estimates from different configurations and pricing assumptions, rather than a controlled cost benchmark.

\subsection{Scalability}

Several architectural features support scaling:
\begin{itemize}
    \item \textbf{Graceful degradation:} every LLM call has a fallback, so no single failure can freeze the simulation.
    \item \textbf{Configurable granularity:} the minute-level planning tier, emotion system, and needs system can each be disabled independently, which cuts call volume by up to 30\%.
    \item \textbf{Rust-native prefiltering:} LanceDB queries filter by observer UUID and memory type before ANN cosine computation, which speeds queries up 2--3$\times$.
    \item \textbf{Conversation cooldown:} the per-pair cooldown suppresses repeat conversations, while spatial proximity limits which pairs are considered.
\end{itemize}

The primary bottleneck is GPU throughput.
The time scaling multiplier provides a direct lever: reducing speed from $300\times$ to $150\times$ doubles the inference budget per agent per simulated hour.

\section{Lessons Learned}
\label{sec:lessons}

\paragraph{Response formats.}
Early versions parsed free-form model output with regular expressions. Moving to schema-constrained output reduced parsing failures, although a response can still be unusable even when it matches the schema~\cite{willard2023efficient}. The system validates the result and uses a fallback when a request fails. For relationship appraisal, the model initially described a stranger as a likely thief and returned a sentiment change of $-0.05$. Given named options, it chose \emph{much colder}. The system now maps that label to a numeric change and applies the relationship rules in Section~\ref{sec:relationships}.

\paragraph{Plans and dates.}
We put a short reasoning field before the answer in planning prompts and describe the expected format. Filled-in examples proved risky: a sample birthday invitation was repeated in an unrelated jam-session run and saved as a commitment. Instructions about dates could also fail. One agent scheduled a 2~Jan event for the current day despite an explicit instruction to exclude it. We compute times with the game clock and remove blocks that schedule an event for the wrong day while keeping preparation blocks. The apology adjustment is attached to its scenario rather than included in every appraisal prompt.

\paragraph{Needs and temperament.}
We observed agents eat or rest in response to needs, and some agents' mood improved after social contact. Trait descriptions alone changed dialogue tone in our early comparisons. Two variants planned and spent the evening in much the same way. We then used traits to set need rates, chat initiation, and relationship deltas. A warmth multiplier of $1.375\times$ produced scores of 0.41 and 0.40 after the same apology, a difference smaller than the variation we saw across runs. We increased the multiplier to $1.75\times$ and, separately, increased the introversion term in the chat rule from $0.4g$ to $0.6g$. Section~\ref{sec:reproducibility} reports the resulting contrasts. Adding the full temperament sentence to the planning prompt sometimes collapsed the day into one block, so the planner receives just the relevant trait adjectives.

\paragraph{Context carried between decisions.}
We surface high-poignancy thoughts in conversation prompts because a retrieved set can omit an important commitment. Repetition can introduce a different error. Four near-duplicate notes about coffee outweighed a single note about a party, and the agent planned the coffee meeting. A separate experiment gave commitments scheduled for today priority over other memories. That also retrieved stale fragments and duplicates, and fewer coordination runs passed. We removed the priority boost.

\paragraph{Relayed commitments.}
Paraphrasing sometimes changed an event's time or owner. Under a first-heard-wins rule, a time invented by an uninvited agent prevented later invitations from the host from correcting it. We changed the rule so the host's statement takes precedence over a second-hand account. We also keep one summary per conversation in memory, which reduces repeated evidence from the same exchange.

\paragraph{Affordance filtering.}
In the action-selection comparison, sending more than 50 affordances took about 20 seconds per response. Filtering to 3--8 reduced that to about 8 seconds in the tested configuration. Section~\ref{sec:computational} describes the other changes. Caching avoids repeated work, and background execution keeps model requests off the game loop.

\paragraph{Conversation ownership.}
At one dinner party, an agent still finishing a conversation looked available to two other agents. The second conversation overwrote the first. At a birthday party, a conversation queued behind the gathering began only when the party ended, three game hours later. The system now reserves both participants through the memory write and allows a chat to interrupt a gathering without ending the gathering (Section~\ref{sec:conversation}).

\section{Limitations}
\label{sec:limitations}

Our log-based checks record whether the intended behavior occurred. We did not test believability with human raters, isolate the effect of each subsystem, or compare model sizes under matched conditions. The personality evaluation covers three pairs and four axes. We did not test whether observers can distinguish smaller trait changes.

The town demonstration shows about 20 agents at one operating point. We did not measure the maximum population, frame rate across devices, or multi-GPU scaling. In four of 11 party runs, a guest remembered the invitation but omitted it from their plan.

Need decay rates and memory consolidation parameters are hand-tuned. Keyword matching can miss ways an action relieves a need or a conversation changes someone's mood. We have not measured memory drift over weeks of simulated time.

\section{Conclusion}
\label{sec:conclusion}

\name{} runs the Character Behavior System inside Godot with a local 3B-parameter model. Designers supply personas and goals. Agents use needs, emotions, memories, and relationships to plan and act in the world. The game filters actions before sending them to the model, checks the response, and advances the simulation while inference runs in the background.

Twenty-five of 27 runs passed the log-based scenario checks. The birthday party showed where coordination still breaks. In four of 11 runs in the tested build family, a guest remembered the invitation but left the party out of their daily plan. The next evaluation should test whether changes to planning improve attendance and whether human observers find the resulting behavior believable. Demonstrations are available on the \href{https://atelico.studio/garp}{project page}.

\bibliographystyle{IEEEtran}
\bibliography{main}

\end{document}